\documentclass{article}

\usepackage{arxiv}

\usepackage[utf8]{inputenc} 
\usepackage[T1]{fontenc}    
\usepackage{hyperref}       
\usepackage{url}            
\usepackage{booktabs}       
\usepackage{amsfonts}       
\usepackage{nicefrac}       
\usepackage{microtype}      
\usepackage{lipsum}

\usepackage{times}
\usepackage{epsfig}
\usepackage{graphicx}
\usepackage{amsmath}
\usepackage{amssymb}
\usepackage{textcomp}
\usepackage[document]{ragged2e}

\usepackage{float}

\title{Unsupervised Domain Adaptation by Optical Flow Augmentation in Semantic Segmentation (A proposal)}

\author{Azeez Oluwafemi\\
Carnegie Mellon University Africa\\
Kigali, Rwanda\\
{\tt\small oazeez@andrew.cmu.edu}
}

\begin{document}
\maketitle

\begin{abstract}
 It is expensive to generate real-life image labels and there is a domain gap between real-life and simulated images, hence a model trained on the latter cannot adapt to the former. Solving this can totally eliminate the need for labeling real-life datasets completely. Class balanced self-training is one of the existing techniques that attempt to reduce the domain gap. Moreover, augmenting RGB with flow maps has improved performance in simple semantic segmentation and geometry is preserved across domains. Hence, by augmenting images with dense optical flow map, domain adaptation in semantic segmentation can be improved.
\end{abstract}


\section{Introduction}
\parindent 10mm

 Semantic segmentation involves being able to label every pixel in an image by assigning each pixel to a specific class. The Problem can be poised as a supervised learning one. This means a machine learning model can be trained on a labeled dataset and then tested on an unlabelled one. Labeling dataset is commonly referred to as annotating. 

  Annotating datasets can be very expensive in terms of time and labor. For example annotating cityscapes dataset takes 90 minutes \cite{cordts2016cityscapes}. In an attempt to overcome such limitations, rendered scenes or synthetic datasets were created such as Grand Theft Auto V \cite{richter2016playing}, SYNTHIA \cite{ros2016synthia}, Virtual Kitti\cite{gaidon2016virtual} and VIPER \cite{richter2017playing}. However, there's a large domain gap between synthetic datasets and real datasets based on appearance. Models trained on synthetic datasets seem to perform poorly when tested on real-life datasets as a result. If this gap can be closed or made infinitesimal or negligible, then we can completely bypass collecting real-life datasets and massively annotate synthetics ones and train with them and then adapt them for real-life inference. 

According to Yuhua Chen et al. \cite{inputoutput}, rich geometric information that could be easily and cheaply obtained from synthetic data, such as surface norm, optical flow, depth, etc has been overlooked, Geometric information don\textquotesingle t suffer from domain shifts and there\textquotesingle s a strong correlation between semantics and geometry. The authors then built a deep network that jointly reasons about depth and semantics. This inspired me to consider reasoning about optical flow. Would augmenting appearance dataset with dense optical flow improve domain adaptation in semantic segmentation?

Mean Intersection Union (mIoU) is the metric used for measuring how well a good semantic segmentation is \cite{csurka2013good}. IoU is estimated for each class and then the mean is taken over all the available classes. If augmentation by optical flow increases the IoU of motion sensitive class, then it would increase the mIoU for the whole task thereby reducing the overall domain gap between synthetic and real data. 

I would be working with adapting a model trained on a synthetic scene  GTA5 \cite{richter2016playing} to a real-life dataset cityscapes \cite{cordts2016cityscapes}. The training technique I plan to use which is unsupervised is class-balance self-training \cite{zou2018unsupervised}. I would then use mIoU \cite{csurka2013good} as a measurement metric. Optical flow maps would be generated with Flownetv2 \cite{ilg2017flownet}. A better mIoU than the one obtained by Zou, Yang et al. \cite{zou2018unsupervised}, would prove that augmenting with optical flow actually improved domain adaptation since the images there were not augmented at all.

This would extend the previous work done by Zou, Yang et al. \cite{zou2018unsupervised} and Yuhua Chen et al. \cite{inputoutput}. Since what I\textquotesingle m doing differently from Zou, Yang et al. \cite{zou2018unsupervised} are just augmenting the images with optical flow based on the inspiration of using depth with a completely different architecture by  Yuhua Chen et al. \cite{inputoutput}. I would also be generating a new idea since it\textquotesingle s the first time images would be augmented with flow maps for the purpose of domain adaptation thereby giving an incite to the role of optical flow maps in helping to solve the domain gap problem.

The approach would be to generate flow maps using Flownet v2 or Open CV Farneback function or ground truth (if available). The existing RGB image would then be augmented with the generated flow maps and finally trained with class balanced self-training. The datasets used would be GTA5 and cityscapes in which 19 classes would be evaluated. The 19 classes include Road, sidewalk, building, wall, fence, pole, traffic light, traffic sign, vegetation, terrain, sky, person, rider, car, truck, bus, train, motorcycle and bike. The GTA5 dataset has 24966 images of size 1052 by 1914. 


\section{Literature Review}
There's been an advance in deep network architectures used for semantic segmentation, one of such is Deeplab v2 \cite{chen2018deeplab}. Input images were passed through a layer of atrous(dilated) convolutional network which helps with adjusting the field of view and control the resolution of the feature map generated by a deep convolutional neural network. A coarse score map is then produced which undergoes bilinear interpolation for upsampling, finally Fully connected conditional random field (CRF) is used as a post-processing process which helps incorporated low-level details in the segmentation result since skip connections are not used here. In an attempt to close domain gaps, the main idea has been to reduce the gap between source and target distribution by learning embeddings that are invariant to domains such as in Deep Adaptation Network (DAN) architecture which generalizes simple convolutional network tasks to fit for domain adaptation \cite{long2015learning}. The base convolutional neural network for Deeplab v2 is Resnet38 \cite{wu2019wider} which has been pretrained on Imagenet dataset \cite{deng2009imagenet} through transfer learning technique\cite{raina2007self}.

One of the first adversarial approaches in dealing with unsupervised domain adaptation was CyCADA (cycle consistent adversarial domain adaptation) \cite{hoffman2017cycada} which combined the concept of cycle consistency and adversarial domain adaptation. In cycle consistency, the goal is to learn the mapping. $ G: X -> Y $, where X is the source domain and Y is target domain. Training $ G(X) \approx Y $ using adversarial loss.  Since this is underconstrained, they introduced cycle consistency loss to enforce $ F(G(X)) \approx X $ by also learning the mapping $ F: Y -> X $ \cite{zhu2017unpaired}. In CyCADA, the goal was to train a target domain encoder by trying to discriminate from a representation of a pretrained source domain. And then predicting with the target domain using the source classifier and avoid divergence by enforcing cycle consistency \cite{hoffman2017cycada}. 

Yuhua Chen et al.
\cite{inputoutput} took this idea further by trying to take advantage of the correlation between semantics and geometry specifically using depth maps. They employed input level adaptation and output level adaptation. They tried to generate a realistic target image using a generator $ G_{img} $  with a combination of synthetic image, synthetic label and synthetic depth map as input and a discriminator $ D_{img} $  which takes target image and the output of the generator as an input to predict fake or real. The output generator is then used for a task of generating labels and depth maps using a generator and discriminator network alike. The intuition for output adaptation is that the network would produce good semantic labels if it tries to produce good depth maps simultaneously since this would help ensure the correlation between semantics and geometry is exploited. 

Zou Yang et al.
\cite{zou2018unsupervised} then applied the concept of class-balanced self-training. A self-training method is used by generating pseudo labels from the target domain based on high confidence prediction and the network is trained end-to-end with a unified domain and task-specific loss minimization. The concept of “class-balancing” is also introduced which involved normalizing the confidence of each pseudo-label with reference confidence for that particular class. Spatial priors were also introduced since traffic scenes share similar spatial priors. This improves cross-domain adaptation and yielded state-of-the-art results. 

Rashed et al.
\cite{rashed2019optical} was able to show that augmenting color images with optical flow maps can improve semantic segmentation. It could improve IoU for moving object classes. They augmented RGB data with optical flow maps using 4 various CNN architectures. Three of them were one stream networks based on FCN8s \cite{long2015fully} and the third one was a two-stream network which was influenced by the works of Simonyan and Zisserman \cite{simonyan2014two} and Modenet by Siam et al \cite{siam2017modnet} They used three various options of depth flows, which are: Colour wheel representation in 3D, Magnitude and direction in 2D and Magnitude in 1D. The Optical flow was then normalized to be in the range 0-255. OpenCV Farneback function was used to generate optical flow magnitude along with FlowNet v2 \cite{ilg2017flownet}. Their results showed a generally better performance for the two-stream RGB and  Flow networks. With a 3 color wheel 3D representation for the flow map.



\section{Proposed Methodology}
The general approach would be to generate a flow map and also use the available ground truth flow map. Augment RGB image with optical flow maps and then use class balanced domain adaptation for domain adaptation.

\subsection{Optical Flow Generator}
\cite{ilg2017flownet} was created for the sole purpose of generating flow maps. Since cityscape dataset does not have a ground truth flow map. There are three options of flow maps representation. They are Colour wheel representation in three dimensions,  magnitude and direction representation in two dimensions and magnitude in one dimension. \cite{rashed2019optical} observed a better performance from the color wheel representation while trying to augment data with optical flow for semantic segmentation, so that would be the representation we would be choosing since our task is similar. Then the optical flow values would be normalized to the range 0 to 255 so that the scale is similar to the equivalent RGB image. Another option that would be explored for generating optical flow maps would be to use OpenCV Farneback function. 

\subsection{Augmentation Approach}
The next stage would now be to augment RGB dataset with a flow map. \cite{ilg2017flownet} also tried augmenting using four CNN architectures. One had an RGB image with RGB encoder, Flow map with Flow encoder, concatenated RGB and flow maps and a two-way stream of RGB image with encoder and Flow map with its encoder. They got their best performance with RGB + F two-way stream. Concatenated RGBF would be the approach we would use since it performs well on a simple segmentation task.

\subsection{Class Balanced Self Training}
Finally, a class balanced self-training method is used for adapting a model trained on the GTA5 dataset to cityscapes. This architecture is used to train a hard case of domain adaptation where the source domain dataset is synthetic and labeled and the target domain is unlabelled and the task is to predict the label for that target dataset. Class Balanced self-training is actually introduced by \cite{zou2018unsupervised}.
 A self-training method is used by generating pseudo labels from the target domain based on high confidence prediction and the network is trained end-to-end with a unified domain and task-specific loss minimization. The concept of “class-balancing” is also introduced which involved normalizing the confidence of each pseudo-label with reference confidence for that particular class.
Spatial priors were also introduced since traffic scenes share similar spatial priors. This would improve cross-domain adaptation.

\subsection{Datasets, Framework, and Base Deep Neural Network}
The datasets used would be GTA5 (synthetic and labeled) and cityscapes (real and unlabelled) in which 19 classes would be evaluated. The GTA5 dataset has 24966 images of size 1052 by 1914. Pytorch, a python framework for building deep neural networks would be used. Deeplab v2 \cite{chen2018deeplab} which had been pre-trained on imagenet \cite{deng2009imagenet} and would be used here implementing transfer learning technique \cite{raina2007self}

\begin{figure}[H]
\begin{center}
\includegraphics[width=1\linewidth]
                  {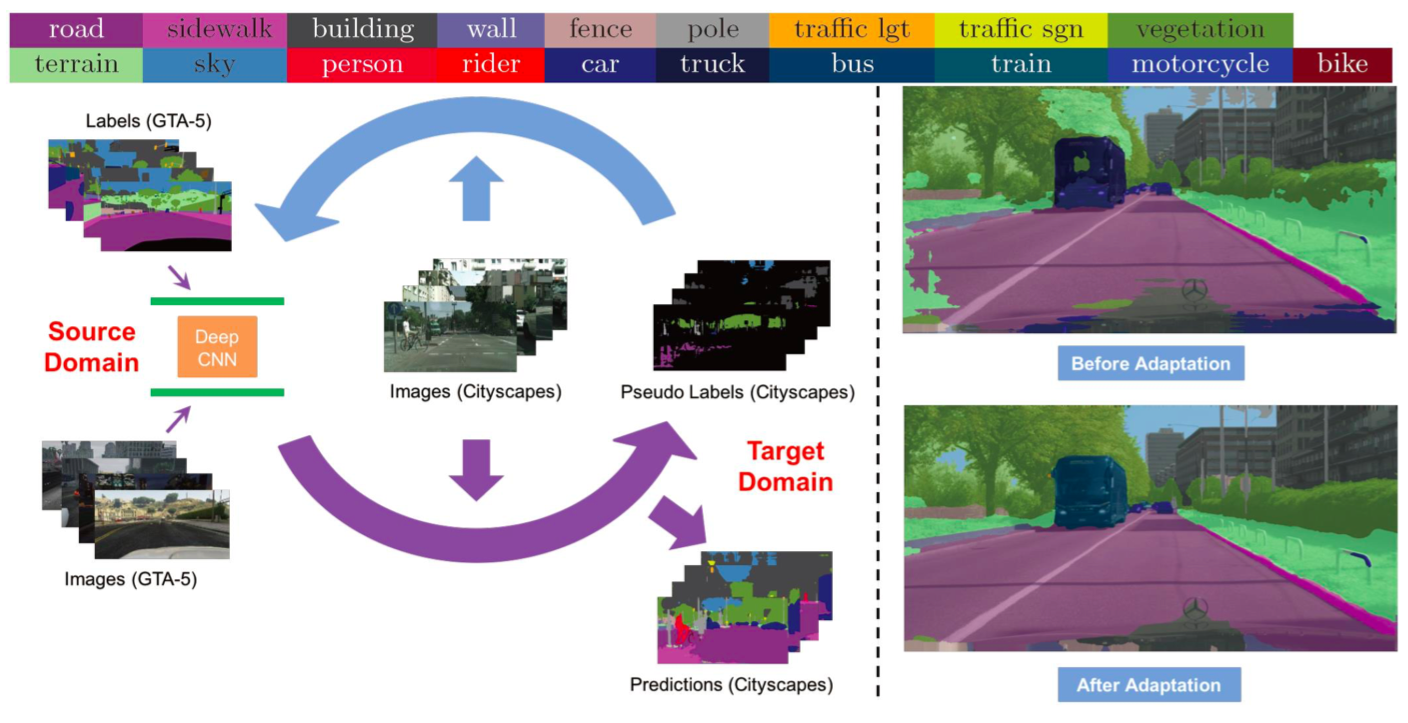}
\end{center}
  \caption{Class balanced self-training.\cite{zou2018unsupervised}}
\label{fig:short}
\end{figure}

\begin{figure}[H]
\begin{center}
\includegraphics[width=1\linewidth]
                  {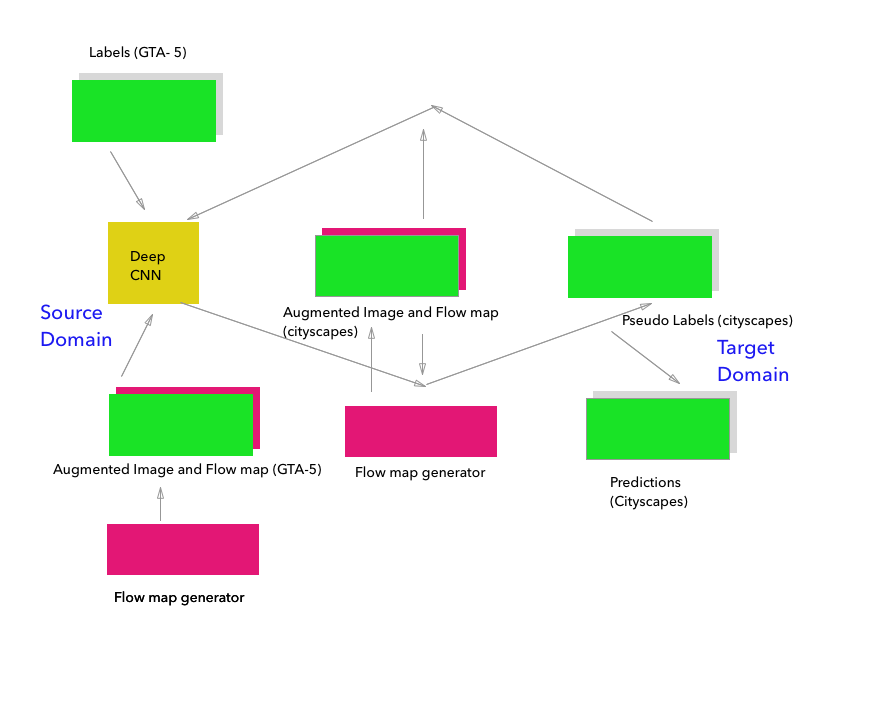}
\end{center}
  \caption{Training process}
\label{fig:short}
\end{figure}

\section{Expected Results}

Mean Intersection Union (mIoU) \cite{csurka2013good} is the metric that would be used for measuring performance. The expectation is that augmenting with the optical flow should increase the IoU of motion sensitive class, then it would increase the mIoU for the whole task thereby reducing the overall domain gap between synthetic and real data. \cite{rashed2019optical} got an increase in IOU of 17 \% and 7 \% on the motorcycle and train classes respectively by augmenting with visual flow map on cityscapes for a semantic segmentation task. The mIoU expected should be greater than 47 \% for the task of adapting from GTA5 to cityscapes. 47 \% is the baseline obtained by \cite{zou2018unsupervised} for this task without augmentation.

\section{Conclusion}
Annotating a real-life image dataset like cityscapes for the task of semantic segmentation can be very expensive, while it is easier to annotate a synthetic dataset like GTA5. Therefore a naive approach of training a model to learn semantic segmentation on synthetic dataset and test on real-life dataset easily comes to mind. It doesn't work easily because of the challenge of domain gap. There's a domain gap in the RGB of both datasets. The reasonable approach is then to try out unsupervised methods of domain adaptation using techniques like class balanced self-training (CBST) \cite{zou2018unsupervised} which produces some results but does not totally close the domain gap. Augmentation of RGB dataset with optical flows is however proposed because the geometry is preserved across both domain, also \cite{rashed2019optical} already was able to perform better at ordinary segmentation task by augmenting RGB with optical flows. A future work could be correcting for doppler effect by adding depth information. 

{\small
\bibliographystyle{unsrt}
\bibliography{output}
}


\end{document}